\documentclass[10pt,journal,compsoc]{IEEEtran}
\ifCLASSOPTIONcompsoc
  \usepackage[nocompress]{cite}
\else
  \usepackage{cite}
\fi
\usepackage{amssymb}
\usepackage{mathdots}
\usepackage{bm}
\usepackage{amsmath}
\usepackage{overpic}
\usepackage{tabularx}
\usepackage{multirow}
\usepackage{subfigure}
\usepackage{threeparttable}
\usepackage{graphicx}
\usepackage{cleveref}
\usepackage{caption}
\usepackage{epsfig}
\usepackage{epstopdf}
\usepackage{color}
\ifCLASSINFOpdf
\else
\fi
\begin{document}
%
\title{Graph Edge Representation via Tensor Product Graph Convolutional Representation}
%
%
%
%

\author{Bo~Jiang,
        Sheng~Ge,
        Ziyan~Zhang,
        Beibei~Wang,
        Jin~Tang
        and~Bin~Luo
\IEEEcompsocitemizethanks{\IEEEcompsocthanksitem 

The authors are with Key Laboratory of Multimodal Cognitive Computation, School of Computer Science and Technology, Anhui University, Hefei 230601, China. E-mail: jiangbo@ahu.edu.cn \\
\protect\\
}
\thanks{Manuscript received April 19, 2005; revised August 26, 2015.}}

%
%

\markboth{Journal of \LaTeX\ Class Files,~Vol.~14, No.~8, August~2015}%
{Shell \MakeLowercase{\textit{et al.}}: Bare Demo of IEEEtran.cls for Computer Society Journals}
%



\IEEEtitleabstractindextext{%
\begin{abstract}
 Graph Convolutional  Networks (GCNs) have been widely studied.
 The core of GCNs is the definition of convolution operators on graphs.
 However, existing Graph Convolution (GC) operators are mainly defined on adjacency matrix and \textbf{ node features} and generally focus on obtaining effective node embeddings which cannot be utilized to address the graphs with (high-dimensional) edge features.
To address this problem, by leveraging tensor contraction representation  and tensor product graph diffusion theories,
this paper analogously defines an effective convolution operator on graphs with \textbf{edge features} which is named as Tensor Product Graph Convolution (TPGC).
The proposed TPGC  aims to obtain effective edge embeddings. It provides a complementary model to traditional graph convolutions (GCs) to address the more general graph data analysis with both node and edge features.
Experimental results on several graph learning tasks demonstrate the effectiveness of the proposed TPGC.
\end{abstract}

}

\maketitle

\IEEEdisplaynontitleabstractindextext

%
\IEEEpeerreviewmaketitle

\IEEEraisesectionheading{\section{Introduction}\label{sec:introduction}}

%
%
%
%
\IEEEPARstart{M}{any} problems in machine learning field can be formulated as representation learning on various kinds of graph type data. 
Recently, Graph Convolutional Networks (GCNs) have been commonly studied for graph data representation and learning~\cite{kipf2017semi,ZHOU202057}.
The core of GCNs is the definition of convolution function on graphs. Existing graph convolution functions can be roughly divided into two groups, i.e., spectral convolution and spatial convolution~\cite{ZHOU202057}.
Spectral convolution commonly defines Graph Convolution (GC) function by
leveraging spectral graph theory~\cite{kipf2017semi,xu2018graph} while spatial methods generally define GCs by  employing some graph propagation/spatial transformation functions~\cite{hamilton2017inductive,velickovic2019deep,jiang2019data,chien2021adaptive,zhu2020bilinear}. 
For instance,
Kipf et al.~\cite{kipf2017semi} propose a simple GCN which is one of most popular GNNs for graph learning tasks.
Xu et al.~\cite{xu2018graph} propose to leverage graph wavelet transform to implement graph convolution.
Liu et al.~\cite{Liu2021Nodewise} propose a node-wise localization of GNNs by considering both global and local information.
Chien et al.~\cite{chien2021adaptive} propose GPR-GNN by introducing Generalized PageRank techniques to combine with GNNs.

\textbf{Related Works. }
However, the above proposed graph convolution functions are generally defined on  adjacency matrix and node features and focus on obtaining effective  embeddings for graph nodes.
Therefore, they  cannot be utilized to address the graphs with (high-dimensional) edge features.
To deal with graph edge features,
\emph{\textbf{one}} simple way is to use an aggregation strategy which
first defines multiple graphs based on different edge feature dimensions and then aggregates
the representations of multiple graphs together for final representation~\cite{yun2019graph,wang2021heterogeneous,mGARL}.
Obviously, these strategies can not obtain embeddings for graph edges and also lack of considering the  dependence of different feature dimensions.
The \emph{\textbf{second}}  popular approach to deal with edge features is to employ an attention architecture~\cite{velickovic2018graph,ye2021sgat,k2018attentionbased,jiang2020coembedding}
which first obtains an attention/weight for each edge
via a mapping function and then performs graph convolution
operation based on the learned weighted graph. 
However, the learning of edges in these works 
are computed solely based on the content features of the edges which thus 
fail to fully model the adjacency
relationships among different edges in edge feature learning.  
The \emph{\textbf{third}} simple way to address edge features is to first construct a {line} graph $L(G)$ to represent the adjacencies between edges and %
then conduct traditional graph convolution operations on line graph to obtain edge embeddings~\cite{Linegraph,Chen2019GL2vec,jiang2020coembedding}.
Note that the graph convolution learning  on line graph $L(G)$ is usually impractical for dense graph $G$ due to
high storage and computing cost.
After revisiting the above related works, it is natural to raise a new question: \emph{can we analogously define a GC-like convolution operator
 for the graphs with edge features directly ?}

\textbf{Contributions. }
In this paper, inspired by tensor contraction representation and tensor product graph propagation theories~\cite{kossaifi2020tensor,yang2012affinity}, we propose a novel  convolution
operation, called  Tensor Product Graph Convolution (TPGC), for graph edge representation and learning. 
Analogical to standard  Graph Convolutions (GCs) defined on adjacency matrix and {node features} for obtaining \textbf{node embeddings},
TPGC is defined on {edge features} and focuses on obtaining effective \textbf{edge embeddings}.
Thus, it provides a complementary model to traditional GCs. 
Comparing with previous  aggregation or attention based edge learning methods, the proposed TPGC
can explicitly obtain context-aware embeddings for graph edges by fully exploiting the structural information of different edges.
Also, contrary to line graph based method, TPGC is defined on original graph and thus needs obviously lower storage and computing cost.
Using the proposed TPGC and GC, we then propose a unified network architecture, named Edge Tensor-Graph Convolutional Network
(ET-GCN) and Edge Tensor-Graph Attention Network
(ET-GAT), for the general graph learning problem with both node and edge features.
%

Experimental results  demonstrate the effectiveness of  TPGC on several graph learning tasks including node classification, link prediction and multi-graph learning.

\section{Preliminaries}

\subsection{Notations}

We present the main mathematical notations for a graph with both node and edge features in the following list.
\begin{itemize}
\item We assume an undirected graph $G$ with node set $V$ and edge set $E$, $E \subseteq V \times V$ where $n = \left|V\right|$ denotes the number of nodes.
\item Let $A \in \mathbb{R}^{n \times n}$ be the adjacency matrix of $G$ in which  each element $A_{ij}$ indicates the connectivity between node $v_i$ and node $v_j$.
\item Let $H \in \mathbb{R}^{n \times d}$ denotes the collection of graph node's features where each node is represented as a $d$-dimensional vector.
\item Let $\mathcal{S} \in \mathbb{R}^{n \times n\times p}$ denotes the collection of graph edge's features where each edge is represented as a $p$-dimensional vector.
\end{itemize}

\subsection{Revisiting GC from Dual Linear Transformations}


We briefly review the widely used graph convolutional network (GCN) proposed in work~\cite{kipf2017semi}.
The core aspect of GCNs is the definition of Graph Convolution (GC) function for node's feature representation in GCN's layer-wise message passing.
Given a graph $G=(H,A)$ with node features $H  \in \mathbb{R}^{n\times d}$  and adjacency matrix ${A}\in \mathbb{R}^{n\times n}$, Kipf and Welling~\cite{kipf2017semi} propose to define GC function as 
%
\begin{equation}
H'\leftarrow \widetilde A HW \label{con:AHW}
\end{equation}
where $W\in \mathbb{R}^{d\times d'}$ denotes the learnable graph convolutional parameters and  $\widetilde A = \bar D^{-\frac{1}{2}}\bar A\bar D^{-\frac{1}{2}}$ denotes
the re-normalization graph by adding self-loop to adjacency matrix ${A}$, i.e., $\bar A = A + I$ and $\bar D$ is the diagonal  matrix with $\bar D_{ii}=\sum_j \bar A_{ij}$, as introduced in work~\cite{kipf2017semi}.
Note that, a non-linear activation function $\sigma(\cdot)$, such as ReLU, SoftMax, is further required after graph convolution operation in each hidden layer of GCN~\cite{kipf2017semi}. 
%
\begin{figure}[!htb]
	\centering
\includegraphics[width=0.435\textwidth]{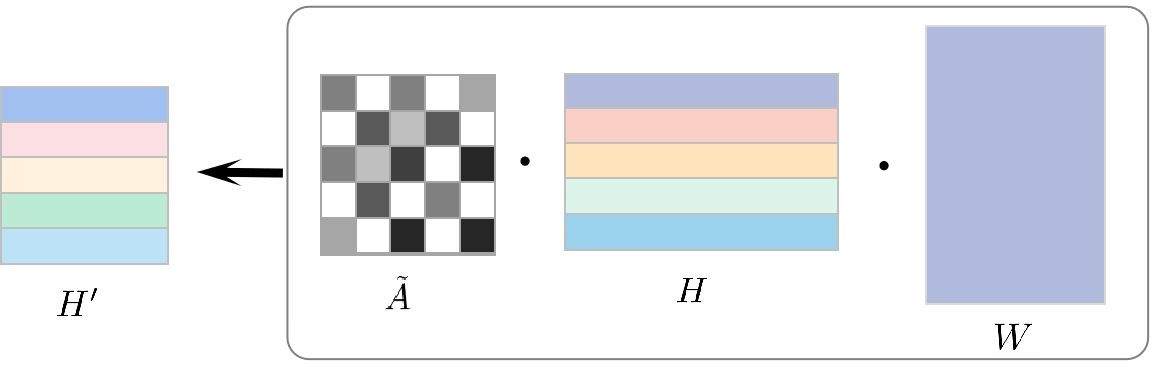}
	\caption{Illustration of GC operation.}\label{fig:AHW}
\end{figure}
%

Figure~\ref{fig:AHW} intuitively shows the process of the above GC function (Eq.(1)). From the view of data representation,
the matrices $\widetilde{A}$ and $W$ in the above GC function act as the \textbf{two-sided linear transformations}~\cite{ye2005generalized,gmlp} on the node's feature matrix $H \in\mathbb{R}^{n\times d}$
to obtain more compact representations $H' \in \mathbb{R}^{n\times d'}$.
Concretely, the {left} side transformation is conducted on \emph{sample} mode/dimension by using the fixed (normalized) adjacency matrix $\widetilde{A}$ for feature propagation, while the {right} side is conducted on \emph{feature} mode via learnable weight matrix $W$ for feature projection, as shown in Figure~\ref{fig:AHW}.

Note that, the two-sided linear transformations have been studied for matrix data representation/compression in machine learning field~\cite{ye2005generalized}.
It has also been exploited in neural network learning in recent years~\cite{gmlp,vaswani2017attention}.
 Given an input matrix data $M\in \mathbb{R}^{k\times d}$, the general dual linear transformations for matrix data representation can be generally formulated as
\begin{equation}
M'\leftarrow  L M R \,\,\,\, \mathrm{or} \,\,\,\,  M'\leftarrow  L \sigma(M R) \label{con:AHW}
\end{equation}
where $L, R$
are learnable weight matrices with proper sizes and $\sigma$ denotes an  activation function~\cite{gmlp}.

 \textbf{Observation:} Here, we can observe that,\emph{ the above GC function can be viewed as a special  case of Eq.(2) in which the two-sided linear transformations are defined  via the fixed adjacency matrix $\widetilde{A}$ and learnable weight matrix $W$, respectively}. 
This observation provides
the basic motivation
for our TPGC formulation, as discussed below.


\section{TPGC Representation}


The above GC is defined on adjacency matrix $A$ and node features $H$ and only  performs convolution operation on node features which cannot be directly utilized to address the graph data with  high-dimensional edge features.
For graph edge features, it is straightforward to represent them via the tensor data form $\mathcal{S}$ in which each element $\mathcal{S}_{ij\cdot}\in \mathbb{R}^p$ denotes the $p$ dimensional feature vector of edge $e_{ij}$ in graph.
Based on this tensor formulation, we develop  a novel Tensor Product Graph Convolution (TPGC) for graph edge's feature representation by leveraging
 tensor contraction~\cite{kossaifi2020tensor} and graph diffusion models~\cite{yang2012affinity}.
Below, we first introduce the general Tensor Contraction Layer (TCL) model. Then, we present our TPGC formulation in \S 3.2.


\subsection{Tensor Contraction Layer (TCL)}

%
Given a graph $G = (\mathcal{S},A)$ with $\mathcal{S}$ denoting the collection of edge features and $A$ denoting the adjacency matrix.
Different from node's features $H$ in 2D matrix form, the edge's features $\mathcal{S}$ is the 3D tensor form.
Similar to Eq.(2) for matrix data,  tensor contraction~\cite{kossaifi2020tensor}
 has been studied for 3D tensor data representation. Given tensor data $\mathcal{S}$,  the general Tensor Contraction Layer (TCL)~\cite{kossaifi2020tensor} for tensor data representation/compression can be formulated as
\begin{equation}
\mathcal{S}'\leftarrow \mathcal{S}\times_1 {U} \times_2 {V}  \times_3 {W}   \label{con:inventoryflow}
\end{equation}
where  $\times_k$ denotes the $k$-th mode product operation\footnote{Given any tensor $\mathcal{X}\in \mathbb{R}^{I_1\times I_2 \times \cdots \times I_N}$ and matrix $A\in \mathbb{R}^{J\times I_k}$, the $k$-th mode product operation is defined as,
$$
\big(\mathcal{X}\times_k {A}\big)_{i_1\cdots i_{k-1} j i_{k+1}\cdots i_N} = \sum^{I_k}\nolimits_{i_k=1} \mathcal{X}_{i_1i_2\cdots i_N} A_{ji_k}   \label{con:inventoryflow}
$$
}
 and $U\in\mathbb{R}^{n\times c}, V\in\mathbb{R}^{n\times r}$ and $W\in\mathbb{R}^{p\times k}$ are learnable weight matrices  which act as  three-sided linear transformations on the tensor data $\mathcal{S}$.
 The output $\mathcal{S}'\in\mathbb{R}^{c\times r\times k}$ gives the learned representation/compression for input data $\mathcal{S}$.
%
\begin{figure}[!htbp]
	\centering
\includegraphics[width=0.45\textwidth]{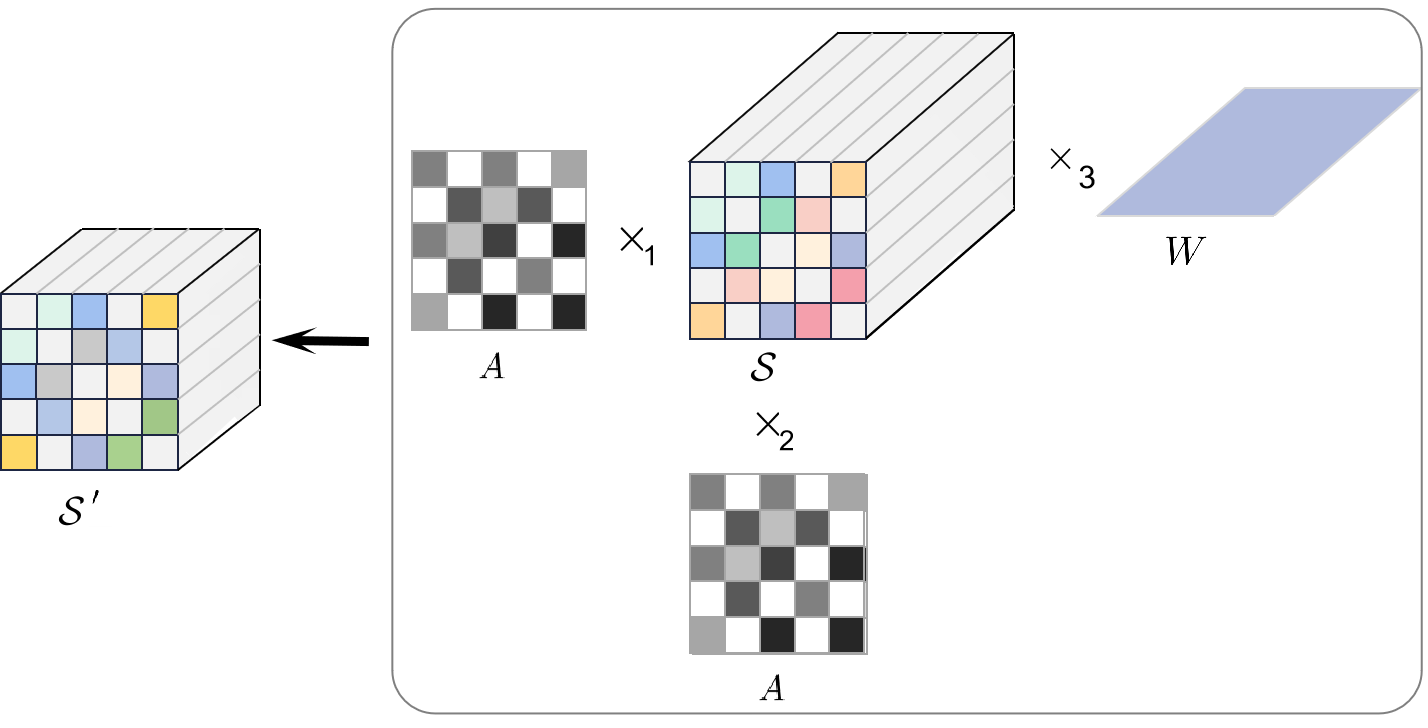}
	\caption{Illustration of the proposed TPGC.}\label{fig:AXW}
\end{figure}
\subsection{From TCL to TPGC}

The above TCL proposes a general formulation for tensor data representation.
In particular, for our edge features $\mathcal{S}$, both the 1st and 2nd modes are {sample} mode and the 3rd mode is the {feature} mode.
Based on the above observation of GC (Eq.(1)),
we can analogically obtain data representation $\mathcal{S}'\in \mathbb{R}^{n\times n\times p'}$ for edge embeddings by employing 1) adjacency matrix ${A}$ for sample mode transformation (propagation) and 2) weight matrix $W\in \mathbb{R}^{p\times p'}$ for feature mode transformation (projection).
Particularly, for sample mode propagation, inspired by work~\cite{yang2012affinity},
we propose to employ a dual propagation mechanism  to conduct propagation on two sample modes of tensor $\mathcal{S}$ simultaneously.
Therefore, we can derive a \emph{special} TCL for data $\mathcal{S}$ by fixing $U=V=\widetilde{A}$ in Eq.(3)  as,
\begin{equation}
\mathcal{S}'\leftarrow \mathcal{S}\times_1 \widetilde{A} \times_2 \widetilde{A}  \times_3 {W}   \label{con:inventoryflow}
\end{equation}
where 
$W\in \mathbb{R}^{p\times p'}$ denotes the weight matrix and $\times_i$ denotes the $i$-th mode product. 
$\widetilde{A}$ denotes the renormalized $A$ which overcomes the scale issue, as defined in Eq.(1). 
The output $\mathcal{S}' \in \mathbb{R}^{n\times n\times p'}$ denotes the learned edge feature representations.
%
In this paper, we call it as Tensor Product Graph Convolution (TPGC) function where $W$ acts as the convolution parameters,  as shown in Figure~\ref{fig:AXW}. 
In order to preserve the self-representation of each edge, we finally propose our TPGC operation as
\begin{equation}
\mathcal{S}'\leftarrow \big( \mathcal{S}\times_1 \widetilde{A} \times_2 \widetilde{A} + \epsilon\mathcal{S}\big)  \times_3 {W}\label{con:inventoryflow}
\end{equation}
where $\epsilon >0$ is a balancing parameter to preserve the self-representation term in graph edge learning. 
%




Using the above TPGC operation,
we can define a multi-layer neural network architecture for graph edge learning.
Specifically, given a graph $G(\mathcal{S},A)$ with $\mathcal{S}^{(0)} = \mathcal{S}$ denoting edge features, we propose to define the layer-wise propagation rule as,
\begin{equation}
\mathcal{S}^{(l+1)}\leftarrow \sigma \Big[\big(\mathcal{S}^{(l)}\times_1  \widetilde{A} \times_2  \widetilde{A} + \epsilon\mathcal{S}^{(l)}\big) \times_3 {W}^{(l)} \Big] \label{con:inventoryflow}
\end{equation}
where $l=0, 1, 2\cdots L$ and $\sigma(\cdot)$ denotes the non-linear activation function, such as ReLU, SoftMax, etc.
Matrix ${W}^{(l)}$ denotes the layer-wise weight matrix.
The outputs $\{\mathcal{S}^{(1)}, \mathcal{S}^{(2)}\cdots \mathcal{S}^{(L+1)}\}$ of the hidden layers give a series of embeddings for graph edges.

In addition, we can further integrate edge attention~\cite{velickovic2018graph} into TPGC and develop Tensor Product Graph Attention (TPGAT) as follows,
\begin{equation}
\mathcal{S}^{(l+1)}\leftarrow \sigma \Big[\big(\mathcal{S}^{(l)}\times_1  \alpha \times_2  \alpha + \epsilon\mathcal{S}^{(l)}\big) \times_3 {W}^{(l)} \Big] \label{con:inventoryflow}
\end{equation}
Here, $\alpha\in \mathbb{R}^{n\times n}$ and $\alpha_{ij}$ denotes the  edge attention~\cite{velickovic2018graph} which is generally learned as,
\begin{align}\label{Eq:gat}
\mathbf{\alpha}_{ij} = \mathrm{Softmax}_{G}\big(f_\Theta({H}_i \| {H}_j)\big)
%
\end{align}
where ${H}_i, {H}_j$ denote the features of nodes $i,j$  and $\|$ denotes the concatenation operation. $f(\cdot)$ denotes a single-layer feedforward neural network parameterized by $\Theta$.

%


\subsection{Complexity Analysis}
In this section, we discuss the main computational complexity of our TPGC.
The above TPGC can be implemented efficiently due to
i) both $\mathcal{S}$ and $A$ are sparse and ii) only elements of $\mathcal{S}'$ associated with edges are computed.
Specifically, Eq.(5) is implemented via two main steps. First, we compute the feature mode product as 
$(\mathcal{S}\times_1 \widetilde{A})_{hjk}=\sum^n_{i=1}\mathcal{S}_{i,j,k}\widetilde{A}_{hi}$. Since $\widetilde{A}$ is sparse and only non-zero elements are computed.  Thus, the complexity of the above 1-th mode product is $\mathcal{O}(|E|np)$ because 
each non-zero element in $\widetilde{A}$ needs 
$np$ calculations and there are $|E|$ number of edges (non-zeros in both $\widetilde{A}$ and $\mathcal{S}$). Similarly, the complexity of the 2-th mode product operation is  $\mathcal{O}(|E|np)$. Finally,  the computation complexity of the 3-th mode product is $\mathcal{O}(|E|pp')$ because each non-zero element in $\mathcal{S}$ has  $pp'$ calculations and there are $|E|$ non-zero elements in $\mathcal{S}$. 
Therefore, the whole computational complexity is $\mathcal{O}\big(|E|(2n+p)p'\big)$.

\subsection{Comparison with Related Works}

Tensor models have been employed in GNNs for higher order information modeling. 
Liu et al.~\cite{liu2020tensor} propose a Tensor Graph Convolutional Networks (TensorGCN) for text classification in which
 TensorGCN~\cite{liu2020tensor} exploits both intra-graph propagation and inter-graph propagations for node and graph level representations respectively.
Note that, both intra and inter graph propagation in TensorGCN are implemented in traditional matrix (dual) transformation forms.
In contrast, our proposed TPGC/TPGAT is defined on tensor mode product and leverages dual tensor diffusion for  edge's message propagation.
In addition, Huang et al.~\cite{huang2020mr} propose Multi-Relational Graph Convolutional Networks (MR-GCN) by exploiting a generalized
tensor product operation.
Xu et al.~\cite{xu2021spatial} present Spatial-Temporal Tensor Graph Convolutional
Network (ST-TGCN) for traffic prediction by deriving a factorized tensor graph convolution.
However, both MR-GCN~\cite{huang2020mr} and ST-TGCN~\cite{xu2021spatial} are proposed for node's representation by exploiting node's tensor representation and edge relational adjacency matrix.
To be specific, MR-GCN employs an eigen-decomposition of Laplacian  tensor to derive a multi-relational graph convolution  and ST-TGCN utilizes a tensor factorized way to define spatial-temporal graph convolution.
In contrast, TPGC/TPGAT is designed to learn edge's representation.
Specifically, it is designed based on Tensor Contraction Layer~\cite{kossaifi2020tensor} and tensor product diffusion model~\cite{yang2012affinity} which adopts a dual diffusion strategy for edge's message propagation on graph. 
In addition, our method is different from 
MPNN~\cite{mpnn2017}. 
MPNN first provides a general framework to formulate many existing GNNs and also proposes to address edge feature learning by first leveraging a neural network model to map edge $e_{ij}$ feature vector to a matrix, i.e., $B(e_{ij})$  and then conducting node’s update via $\sum_{j\in\mathcal{N}_{i}}B(e_{ij})H_j$. Differently, TPGC  aims to learn edge representation via tensor product graph diffusion. It can learn edge feature presentation more effectively by fully capturing the context information of edges.


\begin{figure}[!htbp]
\centering
\includegraphics[scale=0.5]{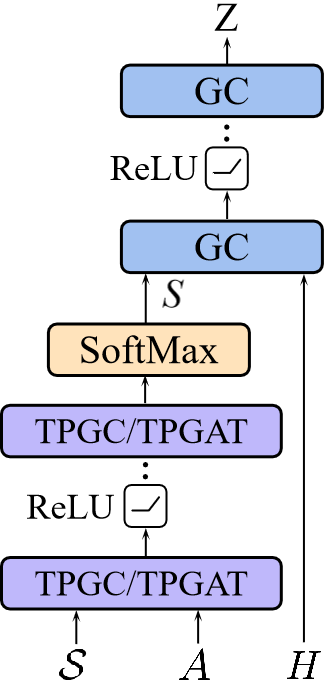}
\caption{The overall framework of ET-GCN and ET-GAT. }
\label{fig:framework}
\end{figure}

\section{Edge Tensor-Graph Convolutional Network}
\begin{table*}[htbp]
		\centering
        \caption{ Comparison results of different methods on three datasets for semi-supervised node classification.}
         \setlength{\tabcolsep}{1.2mm}{
        \label{tab:nodeclassification}
		\begin{tabular}{c|c|c|c|c|c|c|c|c|c|c|c|cccccc}
 \hline
			Data&TrainPCT&GCN&GAT&DGI&DropEdge&R-GCN&MPNN&LNet&AdaLNet&CensNet&ET-GCN&ET-GAT \\\hline
            \multirow{3}{*}{Cora}
            &3\%&79.4$\pm$0.2&80.5$\pm$0.5&79.3$\pm$0.4  &79.4$\pm$0.4
            &78.5$\pm$0.3  &77.1$\pm$1.3
            &76.3$\pm$2.3&77.7$\pm$2.4 &79.4$\pm$1.0  &80.8$\pm$0.7&\textbf{80.9$\pm$0.6}\\

            &1\%&72.9$\pm$0.4&73.6$\pm$0.7 &73.9$\pm$1.4 &72.4$\pm$0.7
            &71.5$\pm$0.2  &67.6$\pm$2.3
            &66.1$\pm$8.2&67.5$\pm$8.7 &67.1$\pm$1.3 &\textbf{75.2$\pm$0.9}&74.8$\pm$0.4\\

            &0.5\%&60.2$\pm$0.6&59.5$\pm$3.4&63.8$\pm$2.1  &57.2$\pm$1.2
            &57.6$\pm$0.2  &54.1$\pm$1.2
            &58.1$\pm$8.2&60.8$\pm$9.0&57.7$\pm$3.9
            &\textbf{64.2$\pm$2.5}&63.9$\pm$1.9\\ \hline

            \multirow{3}{*}{Citeseer}

            &1\%&61.3$\pm$0.5&60.4$\pm$0.8 &62.0$\pm$1.7 &60.2$\pm$1.1
            &59.7$\pm$0.2&60.1$\pm$1.3
            &61.3$\pm$3.9&\textbf{63.3$\pm$1.8}&62.5$\pm$1.5
            &62.6$\pm$0.8&62.3$\pm$0.6\\

            &0.5\%&58.8$\pm$1.7&59.6$\pm$0.7 &56.5$\pm$1.3  &56.2$\pm$3.4
            &58.4$\pm$0.2  &56.0$\pm$1.5
            &53.2$\pm$4.0&53.8$\pm$4.7&57.6$\pm$3.0
            &60.4$\pm$1.2&\textbf{60.6$\pm$0.9}\\

            &0.3\%&47.8$\pm$1.0&49.2$\pm$2.1&49.1$\pm$1.5 &44.6$\pm$2.2
            &48.9$\pm$0.5  &42.5$\pm$1.3
            &44.4$\pm$4.5&46.7$\pm$5.6&49.4$\pm$3.6 &\textbf{50.4$\pm$1.0}&49.8$\pm$1.1\\ \hline

            \multirow{3}{*}{Pubmed}
            &0.1\%&76.2$\pm$0.4&76.3$\pm$0.5 &75.2$\pm$0.8  &73.0$\pm$0.9
            &74.8$\pm$0.1  &72.5$\pm$1.4
            &73.4$\pm$5.1&72.8$\pm$4.6&69.9$\pm$2.1
            &\textbf{76.8$\pm$0.4}&76.7$\pm$0.4\\

            &0.05\%&70.4$\pm$2.4&69.5$\pm$1.1 &68.1$\pm$2.6&66.6$\pm$0.8
            &70.2$\pm$0.1  &66.2$\pm$1.0
            &68.8$\pm$5.6&66.0$\pm$4.5&65.7$\pm$1.2
            &71.2$\pm$1.0&\textbf{71.7$\pm$0.7}\\

            &0.03\%&68.3$\pm$2.5&67.6$\pm$1.5& 57.7$\pm$2.4&63.1$\pm$0.9
            &68.3$\pm$0.2  &63.8$\pm$1.7
            &60.4$\pm$8.6&61.0$\pm$8.7&61.4$\pm$2.8
            &\textbf{68.9$\pm$1.1}&68.5$\pm$1.6\\
            \hline
		\end{tabular}}
\end{table*}

Using TPGC (Eq.(6)) and TPGAT (Eq.(7)),
we can propose a unified network architecture, named Edge Tensor-Graph Convolutional Network (ET-GCN) and Edge Tensor-Graph Attention Network (ET-GAT), for the general graph learning problem with both node and edge features.
The whole overflow of the proposed ET-GCN and ET-GAT is shown in Figure 3 which mainly consists of several TPGC/TPGAT layers and GC layers. 
Let $G(H,\mathcal{S},A)$ be the input general graph in which $H\in\mathbb{R}^{n\times d}$ denotes the node features, $\mathcal{S}\in \mathbb{R}^{n\times n\times p}$ represents  edge features and $A\in \mathbb{R}^{n\times n}$ is the adjacency matrix.
TPGC/TPGAT module aims to generate effective edge's representations and outputs the weighted graph $S\in \mathbb{R}^{n\times n}$ for the followed GC module.
GC is conducted on the weighted graph $S$ which can obtain effective node's embeddings $Z$ for the  downstream tasks, such as semi-supervised node classification, link prediction, etc.
For example, for node classification problem, a softmax function is further employed on node's embeddings $Z$ to obtain the final label prediction. 
%
For link prediction task, the inner product decoder is further used on node's embeddings to obtain the reconstructed adjacency matrix $\hat A$. 
The whole ET-GNNs network are trained in an end-to-end manner via using the ADAM algorithm~\cite{kingma2017adam} which is initialized with Glorot initialization~\cite{Glorot}, as suggested in GCN~\cite{kipf2017semi}.

\section{Experiments}
%

\subsection{Datasets}

We test our method on several benchmark datasets, including three widely used citation networks Cora, Citeseer and Pubmed~\cite{2008Collective} for semi-supervised node classification and link prediction tasks,
two large-scale datasets ogbn-arxiv~\cite{hu2020open} and CIFAR10~\cite{Dang2021DoublyCD,selfsagcn_cvpr21} for semi-supervised node classification,
two multi-graph dataset Handwritten numerals~\cite{AsuncionNewman2007} and Caltech101-7~\cite{AAAI159641}
for multiple graph learning task
and one human protein-protein interaction network PPI~\cite{Zitnik2017,velickovic2018graph} for inductive learning task.

\textbf{Cora} contains $2708$ nodes and $5429$ edges. 
Each node has a $1433$ dimension feature descriptor and all nodes are classified into seven classes.
\noindent\textbf{Citeseer} consists of $3327$ nodes  and $4732$ edges. Each node has a $3703$ dimension feature descriptor. All nodes are classified into six classes.
\noindent\textbf{Pubmed} contains $19717$ nodes and $44338$ edges. 
Each node is represented by a $500$ dimension feature descriptor and all nodes are classified into three classes.
{
\noindent\textbf{ogbn-arxiv} is a large-scale citation network dataset consisting of $169,343$ nodes. 
Each node represents has a $128$ dimension feature descriptor. All nodes are classified into $40$ classes.
\noindent\textbf{CIFAR10} is a collection of $60,000$ color images in $10$ different classes, with $6,000$ images per class. We first extract CNNs feature for every image and then build a k-NN graph for them (k=6).
}
\noindent\textbf{Handwritten numerals} is the digit image dataset consists of $2000$ samples with $10$ different classes from $0$ to $9$.
Following work~\cite{mGARL}, six graphs are constructed by using different views of feature descriptors.
%
\noindent\textbf{Caltech101-7} is an image dataset with $101$ classes. Following work~\cite{mGARL}, we select $7$ categories of them and there are $1474$ images used in our experiments. We also use six kinds of visual feature descriptors to build six neighbor graphs, as utilized in work~\cite{mGARL}. 
\noindent\textbf{PPI} consists of graphs corresponding to different human tissues~\cite{Zitnik2017}. 
It contains $24$ graphs in which the average number of nodes per graph is $2372$. All nodes are classified into $121$ classes.


\subsection{Semi-supervised Node Classification}
\textbf{Experimental setting.}
We compare our method with some other methods on three citation networks including Cora, Citeseer and Pubmed~\cite{2008Collective} and
two large-scale datasets ogbn-arxiv~\cite{hu2020open} and CIFAR10~\cite{selfsagcn_cvpr21,Dang2021DoublyCD}.
To obtain edge features $\mathcal{S}$,
similar to previous works~\cite{velickovic2018graph,gong2019exploiting}, we first use one layer of standard GC~\cite{kipf2017semi} (Eq.(1)) 
to reduce the dimension of each node's feature to $8$ and then use the concatenation of node pairs to build edge features $\mathcal{S}$.
For general three citation datasets, we follow the same experimental setting mentioned in pervious works~\cite{liao2018lanczosnet,jiang2020coembedding}. 
To be specific, for each dataset, we split nodes into three groups, i.e. training, validation and testing.
 We randomly select some nodes for training with three different label rates, i.e., $3\%$, $1\%$, $0.5\%$ for Cora dataset, $1\%$, $0.5\%$, $0.3\%$ for Citeseer dataset and $0.1\%$, $0.05\%$, $0.03\%$ for Pubmed dataset, respectively. Meanwhile, $50\%$ nodes in each dataset are chosen randomly as validation set and the remaining nodes are used for testing.
 As for ogbn-arxiv, we follow the setup from previous work~\cite{hu2020open}. And for image dataset CIFAR10, We randomly select $100$ images in each category as the training set, another $100$ as the validation set, and the remaining images are used for testing.
As shown in Figure 3, in our ET-GNNs, we employ two TPGC/TPGAT layers and set the number of units in hidden layer and output layer to $8$ and $1$ respectively. 
{For node  learning module, we use two GC layers and set the number of units in hidden layer to $128, 32$ according to different datasets. The number of units in output layer is set to $c$ where $c$ denotes class number. The last GC layer outputs the final label predictions for all nodes.
}
The proposed network is trained for a maximum of $10000$ epochs by using the Adam algorithm~\cite{kingma2017adam} with learning rate of $0.01$.
The loss function is defined as the cross entropy classification loss~\cite{kipf2017semi}.
When the validation loss does not decline for $100$ consecutive epochs, the training is stopped.
{The results reported in Table~\ref{tab:nodeclassification} and  Table~\ref{tab:large} are the average performance of $10$ times with different network initializations.}
\begin{table}[!htbp]
		\centering
        \caption{Comparison results of different methods on two large-scale datasets.}
        \label{tab:large}
		\begin{tabular}{c|c|c}
\hline
			Method&ogbn-arxiv&CIFAR10\\
\hline

            GCN&71.86$\pm$0.15& 84.83$\pm$0.04\\
            GraphSAGE&71.86$\pm$0.23&85.22$\pm$0.04\\
            GAT-1 head&70.57$\pm$0.22& 84.92$\pm$0.11\\
            GAT-2 head&71.48$\pm$0.25&84.54$\pm$0.27\\
            DropEdge & 71.06$\pm$0.32&84.59$\pm$0.05\\
            \hline
            ET-GCN &72.01$\pm$0.25&\textbf{85.61$\pm$0.07}\\
            ET-GAT & \textbf{72.13$\pm$0.18}&85.53$\pm$0.07\\
            \hline
		\end{tabular}
\end{table}
\begin{table}[!htbp]
		\centering
        \caption{Comparison results of different methods on PPI dataset for inductive learning task.}
        \label{tab:ppi}
		\begin{tabular}{c|c}
\hline
			Method&PPI\\
\hline

            Random&39.6\\
            MLP&42.2\\
            GraphSAGE-GCN&50.0\\
            GraphSAGE-mean&59.8\\
            GraphSAGE-LSTM&61.2\\
            GraphSAGE-pool&60.0\\
            GAT&97.3\\
            R-GCN&97.0\\
            MPNN&94.4\\
            IGNN&97.6\\
            JK-Net&97.6\\\hline
            ET-GCN &\textbf{98.3$\pm$0.3}\\
            ET-GAT &98.2$\pm$0.2\\
            \hline
		\end{tabular}
\end{table}

\textbf{Comparison results. }
To evaluate the effectiveness of the proposed ET-GNNs (ET-GCN and ET-GAT), we compare it with some other competing edge learning methods which include 1) aggregation strategy based method R-GCN~\cite{schlichtkrull2018modeling}, MPNN~\cite{mpnn2017}, 2) attention architecture GAT~\cite{velickovic2018graph} and 3) line graph based model CensNet~\cite{jiang2020coembedding}.
In addition, we also compare the proposed ET-GCN and ET-GAT with some other GNNs including GCN~\cite{kipf2017semi}, DGI~\cite{velickovic2019deep}, DropEdge~\cite{rong2020dropedge}, LNet~\cite{liao2018lanczosnet} and AdaLNet~\cite{liao2018lanczosnet}. 
DropEdge is a general scheme and we incorporate DropEdge technique in training GCN in experiments.
The comparison results of LNet~\cite{liao2018lanczosnet}, AdaLNet~\cite{liao2018lanczosnet} and CensNet~\cite{jiang2020coembedding} have been reported in their papers and we use them directly.
%
The comparison results are summarized in Table~\ref{tab:nodeclassification} in which the best results are marked as bold.
Here, we can note that
(1) The proposed ET-GNNs generally obtain better performance than traditional GNNs without edge learning module such as GCN~\cite{kipf2017semi},  DGI~\cite{velickovic2019deep}, LNet~\cite{liao2018lanczosnet} and AdaLNet~\cite{liao2018lanczosnet}. This demonstrates the usefulness of the proposed TPGC/TPGAT edge learning module in guiding the learning of graph node's representations.
(2) The proposed ET-GNNs (ET-GCN and ET-GAT) perform better than aggregation based edge learning methods including R-GCN~\cite{schlichtkrull2018modeling} and MPNN~\cite{mpnn2017}  which further demonstrates the more effective of the proposed multi-layer  TPGC  architecture on conducting graph edge learning.
(3) Our ET-GNNs outperform attention based method GAT~\cite{velickovic2018graph}.
(4) The proposed ET-GNNs obtain better performance than line-graph based edge learning  CensNet~\cite{jiang2020coembedding}. This obviously  indicates the more advantage  of the proposed TPGC for edge feature representation than line-graph models. 
{(5) ET-GNNs can achieve better results than the current popular methods on the large-scale dataset ogbn-arxiv and image dataset CIFAR10, which verifies the generality of the proposed ET-GNNs on various dataset tasks and the reliability on large-scale graph datasets.}
(6) Overall, ET-GAT obtains similar performance with ET-GCN on all tasks.

%

\begin{table*}[htbp]
		\centering
        \caption{Comparison results of different methods for link prediction.}
        \label{tab:link}
        \begin{tabular}{c|c|c|c|c|c|c}
         \hline
        	\multirow{2}{*}{Method} & \multicolumn{2}{c|}{Cora}& \multicolumn{2}{c|}{Citeseer}& \multicolumn{2}{c}{Pubmed} \\\cline{2-7}
        	                     &  AUC  &   AP &  AUC  &   AP&  AUC  &   AP\\  \hline
        SC&84.6$\pm$0.01&88.5$\pm$0.00&80.5$\pm$0.01&85.0$\pm$0.01&84.2$\pm$0.02&87.7$\pm$0.01\\
        DW&83.1$\pm$0.01&85.0$\pm$0.00&80.5$\pm$0.02&83.6$\pm$0.01&84.2$\pm$0.00&84.1$\pm$0.00\\
        VGAE&91.4$\pm$0.01&92.6$\pm$0.01&90.8$\pm$0.02&92.0$\pm$0.02&94.4$\pm$0.02&94.7$\pm$0.02\\
        CensNet-VAE&91.7$\pm$0.02&92.6$\pm$0.01&90.6$\pm$0.01&91.6$\pm$0.01&95.5$\pm$0.03&95.9$\pm$0.02\\
        LGLP&92.3$\pm$0.27&93.6$\pm$0.11  &90.9$\pm$0.26&92.7$\pm$0.26 &\textbf{96.9$\pm$0.18}  &\textbf{96.5$\pm$0.41}\\   \hline
        ET-GCN  &\textbf{93.8$\pm$0.21} &\textbf{93.8$\pm$0.30} &92.4$\pm$0.45 &92.9$\pm$0.31   &95.0$\pm$0.04  &95.1$\pm$0.08\\
        ET-GAT  &93.1$\pm$0.93 &93.4$\pm$0.85 &\textbf{93.9$\pm$0.71} &\textbf{94.3$\pm$0.80}   &94.9$\pm$0.15  &95.1$\pm$0.16\\
         \hline
        \end{tabular}

\end{table*}
\subsection{Inductive Learning}

\textbf{Experimental setting.}
For inductive learning task, we evaluate our ET-GNNs  on PPI dataset~\cite{Zitnik2017} which contains $20$ graphs as training set, $2$ for validation set and $2$ for testing set. Specifically, we use the preprocessed dataset version provided by~\cite{hamilton2017inductive} which is commonly used for inductive learning evaluation.
Similar to the above node classification task, the feature representation $\mathcal{X}_{ij}$ of each edge is constructed by conducing subtraction operation on the connected node's features, as used in GAT~\cite{velickovic2018graph}.
We use one TPGC/TPGAT layer and two GC layers to learn the representation of edges and nodes. 
We train ET-GCN and ET-GAT for a maximum of $10000$ epochs  by using the Adam algorithm~\cite{kingma2017adam} with learning rate of $0.005$. The training process is early stopped when the validation loss does not decrease for $100$ consecutive epochs.

\textbf{Comparison results.}
For inductive learning task,  we compare our ET-GCN and ET-GAT method with some other methods including Random, MLP, GraphSAGE~\cite{hamilton2017inductive}, IGNN~\cite{gu2020implicit} and JK-Net~\cite{pmlr-v80-xu18c}.
Also, we compare our ET-GCN and ET-GAT with some edge learning methods including R-GCN~\cite{schlichtkrull2018modeling}, MPNN~\cite{mpnn2017} and GAT~\cite{velickovic2018graph}.
Table~\ref{tab:ppi} summarizes the comparison results on PPI dataset. The results of these comparison methods have been reported in their original papers and we use them directly.
Here, we can note that
(1) ET-GCN and ET-GAT perform better than other popular comparison methods including GraphSAGE~\cite{hamilton2017inductive}, IGNN~\cite{gu2020implicit} and JK-Net~\cite{pmlr-v80-xu18c} which
further demonstrates the effectiveness of the proposed ET-GCN and ET-GAT on conducting inductive graph learning tasks.
(2) ET-GCN and ET-GAT outperform some edge learning methods including GAT~\cite{velickovic2018graph}, R-GCN~\cite{schlichtkrull2018modeling} and MPNN~\cite{mpnn2017} which further demonstrates the advantage and superior performance of the proposed edge learning GNNs. 
The reason why ET-GNN outperforms MPNN is that ET-GNN can learn edge feature presentation more effectively by fully capturing the context information of edges. 
(3) ET-GAT obtains similar performance with ET-GCN on PPI dataset.

\subsection{Link Prediction}
In additional to the above node classification, graph link prediction is also a common  graph learning problem.
The aim of graph link prediction is to learn and predict the existence of potential links in graphs.
We evaluate the effectiveness of the proposed method on link prediction tasks on three citation datasets~\cite{2008Collective}.

\textbf{Experimental setup. }
We follow the experimental setting used in VGAE~\cite{kipf2016variational}. 
We use two evaluation metrics including Area under ROC curve (AUC)
and Average Precision (AP) and report the average result of $10$ runs with random network initializations.
We utilize two TPGC/TPGAT layers and two GC layers to implement ET-GNNs model.
 The hidden units of TPGC layer are set to $\{8, 1\}$. For Cora and Citeseer, the hidden units of GC layers are set to $\{64, 32\}$. For Pubmed, the hidden units of GC layers are set to $\{256, 64\}$. We train   $200$ epochs for Cora and Citeseer, $1000$ epochs for Pubmed by using the Adam algorithm~\cite{kingma2017adam} with learning rate of $0.01$. 

\textbf{Comparison results.}
For link prediction task, we compare the proposed method with some other graph-based   models including Spectral Clustering (SC)~\cite{Lei2011Leveraging} and DeepWalk (DW)~\cite{DeepWalk}, VGAE~\cite{kipf2016variational}, CensNset-VAE~\cite{jiang2020coembedding} and LGLP~\cite{Linegraph}.
Note that, both CensNset-VAE~\cite{jiang2020coembedding} and LGLP~\cite{Linegraph} employ line-graph based edge learning module and thus are related with our work. Table~\ref{tab:link} reports the results of comparison methods.
The results of comparison methods (SC, DW, VGAE and CensNset-VAE) have been reported in work~\cite{jiang2020coembedding} and we use them directly.
Here, we can observe that
(1) The propose method generally obtains  better performance than the competitive line-graph neural  networks including CensNset-VAE~\cite{jiang2020coembedding} and LGLP~\cite{Linegraph}. This clearly demonstrates the advantage of the proposed TPGC module on conducting graph edge learning and thus link prediction tasks.
(2) Our method can obtain  better performance than some other classic link prediction models including Spectral Clustering (SC)~\cite{Lei2011Leveraging}, DeepWalk (DW)~\cite{DeepWalk} and VGAE~\cite{kipf2016variational} which further demonstrates the effectiveness of the proposed edge learning method on this task.
(3) ET-GAT obtains slightly better  performance than ET-GCN on link prediction task.

\begin{figure}[htbp]
\centering
\includegraphics[scale=0.215,trim=110 0 0 0, clip]{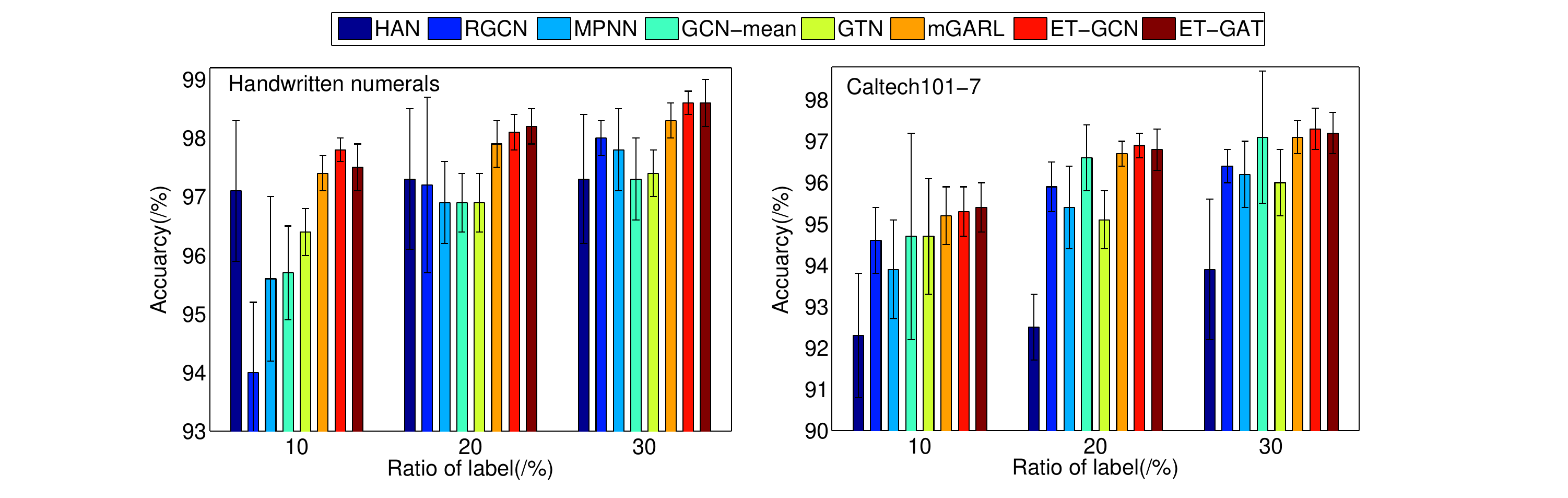}
\caption{Comparison results  for multiple graph learning. }
\label{fig:multigraph}
\end{figure}

\subsection{Multiple graph learning}

In many real-world applications, graph data  come with multiple types of edges which is known as
multi-graph learning problem~\cite{Yu2020MGAT,mGARL}.
Here, we use $G(H,\mathcal{A})$ to denote the input multi-graph data, where $H\in \mathbb{R}^{n\times d}$ denotes $n$ nodes' features and $\mathcal{A}=\{A^{(1)},A^{(2)},\cdots, A^{(m)}\}$ represents  multiple graph structures~\cite{mGARL}.  
Based on these notations, multiple graph representation aims to learn a consistent latent representation $Z = g(H,\mathcal{A};\Theta)$ for the learning tasks that takes in multiple graph structures $\mathcal{A}$ together.
In our experiments, we mainly focus on semi-supervised node classification. 
We use two multi-graph datasets, i.e., Handwritten numerals~\cite{AsuncionNewman2007} and Caltech101-7~\cite{AAAI159641} and conduct semi-supervised node classification for evaluation.


\textbf{Experimental setting.}
To address multiple graph learning with TPGC model,
we first construct edge feature tensor data $\mathcal{S}=\{A^{(1)},\cdots, A^{(m)}\}\in \mathbb{R}^{n\times n\times m}$ by stacking different adjacency matrices together, as suggested in work~\cite{huang2020mr} and then use the proposed ET-GNNs to learn the node's embeddings for classification.
For all datasets, we randomly select $10\%$, $20\%$ and $30\%$ labeled nodes in each class for training. Besides, we select another $5\%$ for validation and the rest nodes for testing.
The results reported in Figure~\ref{fig:multigraph} are the average of $10$ times with different data splits.
We use two TPGC/TPGAT layers and two GC layers to learn the edge and node representations respectively. The numbers of hidden units are set to $\{6, 1\}$ and $\{16, c\}$ respectively, where $c$ denotes the number of 
 classes.
The learning rate is $0.005$ 
and the other network's settings are same as that in the above semi-supervised classification task.

\textbf{Comparison results.}
First, we compare the proposed ET-GNNs with simple multiple graph fusion method, i.e. GCN-mean, which uses the averaged adjacency matrix $A=\frac{1}{m}\sum_{v=1}^mA^{(v)}$ as input and then adopts traditional GC for node's embedding and learning. Also, we make comparison with a few methods considering edge feature including R-GCN~\cite{schlichtkrull2018modeling} and MPNN~\cite{mpnn2017}.
 Besides, we compare our method with several recent multi-graph learning methods including GTN~\cite{yun2019graph}, HAN~\cite{wang2021heterogeneous} and mGARL~\cite{mGARL}. Figure \ref{fig:multigraph}  summarizes  the comparison results on two datasets.
 Here, we can note that, the proposed method generally performs better than other  multi-graph learning methods on all cases.
 This further shows the effectiveness of the proposed ET-GCN and ET-GAT to conduct multi-graph learning tasks.

\subsection{Visual Demonstration and Parameter Analysis}



\begin{figure}[htbp]
\centering
\subfigure[Handwritten numerals]{\includegraphics[width=0.25\textwidth]{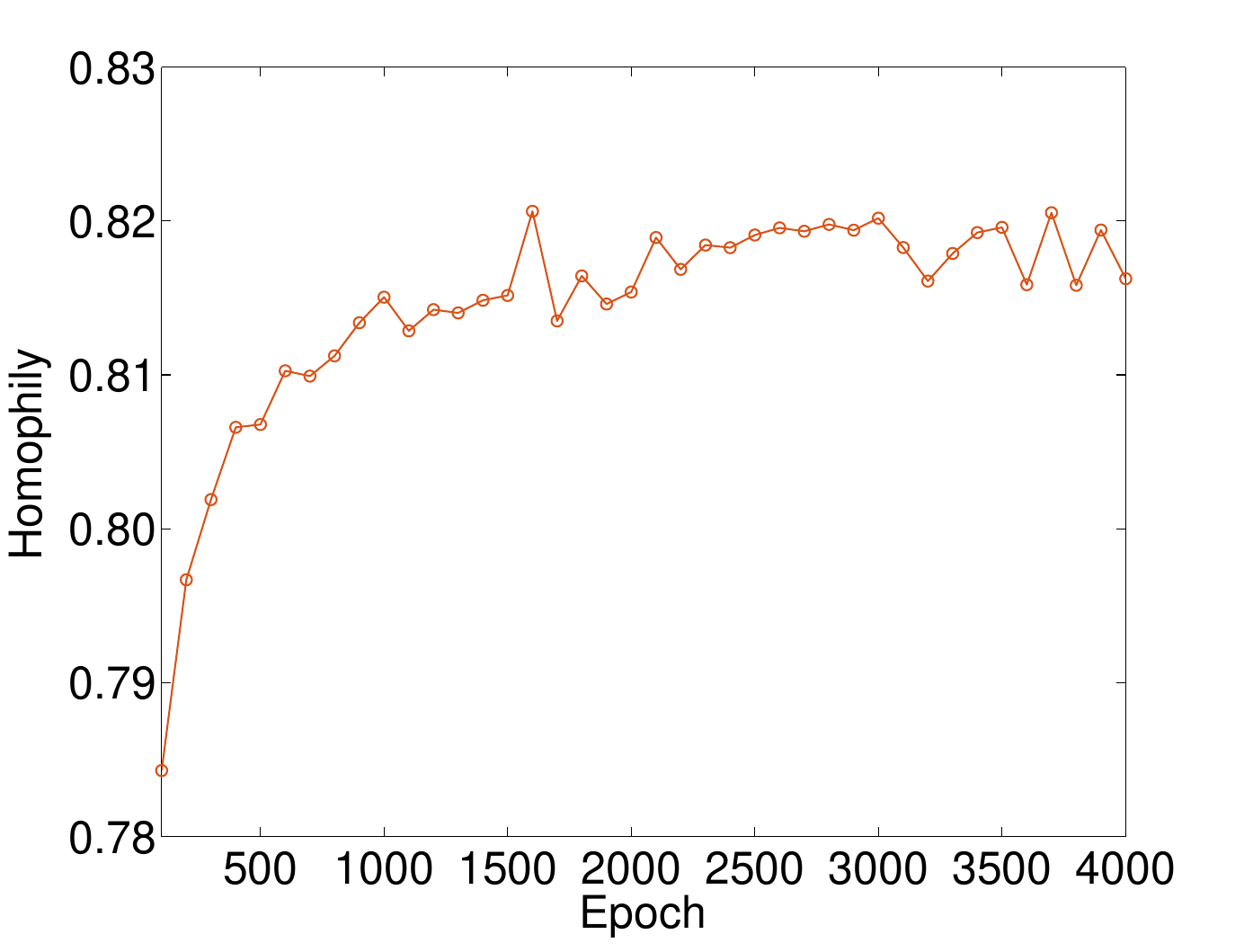}}
\hspace{-4.4mm}
\subfigure[Caltech101-7]{\includegraphics[width=0.25\textwidth]{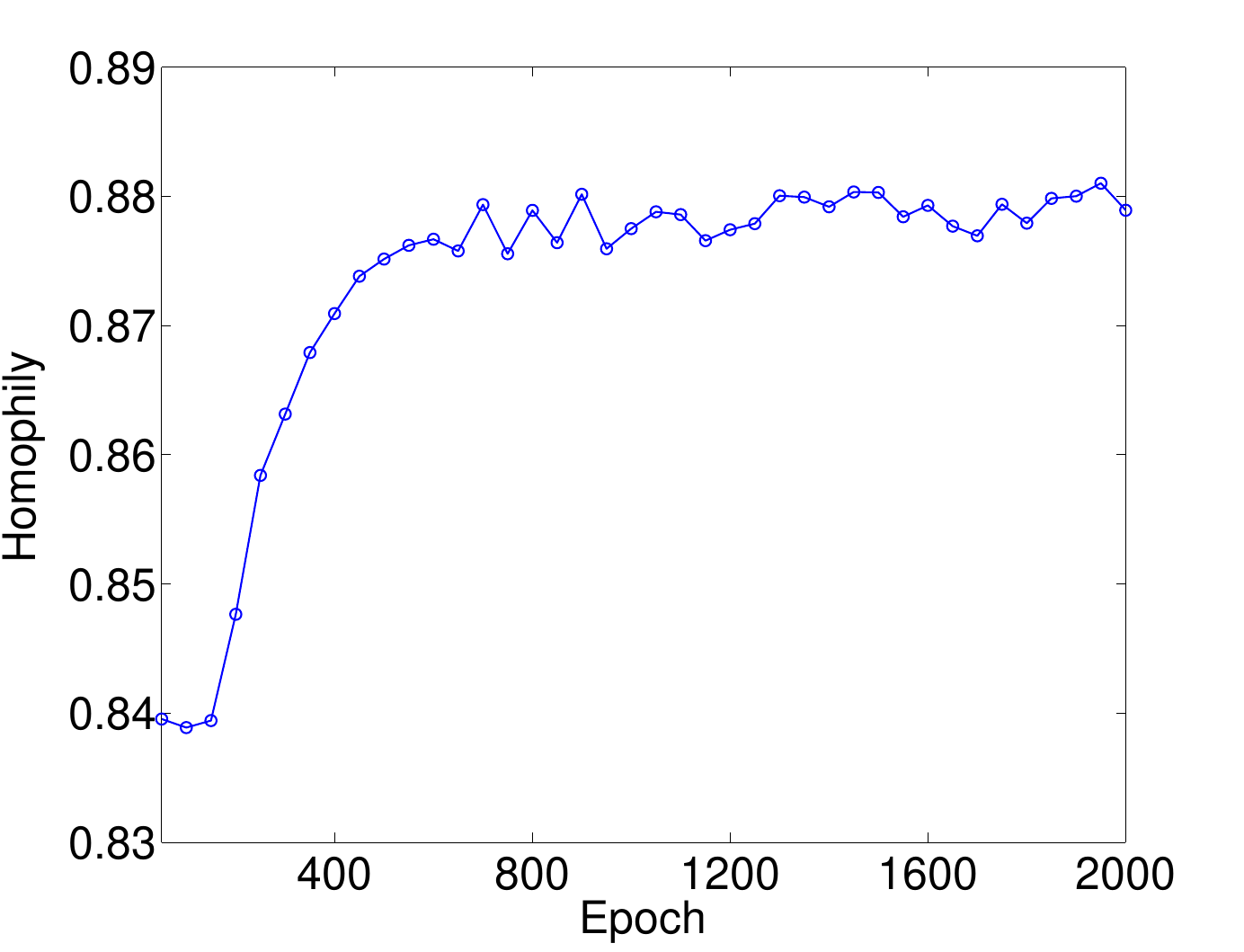}}
 \caption{Comparison results of homophily of the learnable adjacency matrix obtained by TPGC.}
 \label{fig:labelcon}
\end{figure}

In this section, we show the homophily property~\cite{jin2022automated} of the learned $\mathcal{S}$.
We use the label consistency, i.e., the proportion/rate of neighboring nodes that have the same label with target node, as defined in work~\cite{labelcon2020}.
Figure~\ref{fig:labelcon} demonstrates the homophily of the learned $\mathcal{S}$ across different training epoches
%
on Caltech101-7~\cite{AAAI159641} and Handwritten numerals~\cite{AsuncionNewman2007} datasets. 
 As the training epoch increases, the learned graph $\mathcal{S}$ contains more and more links (edges) between graph nodes that have the same class label. Intuitively, this property is beneficial for the semi-supervised classification task.
Furthermore, we investigate the performance of ET-GCN under different settings of TPGC module.
We set the depth of TPGC-layer to 2 and the number of hidden units is set to $8$ to obtain the best average learning performance.
Table \ref{tab:depth} shows the performance of ET-GCN with different depths of TPGC layer on Cora dataset.
Table \ref{tab:hidden} further demonstrates the performance of ET-GCN with different numbers of hidden units of TPGC layer on Cora dataset.
Note that, ET-GCN maintains better performance under different depths of TPGC layer, which generally indicates the insensitivity of
the proposed ET-GCN w.r.t model depth.
Also, one can note that ET-GCN performs insensitively w.r.t number of hidden units. 
%
{In addition, to evaluate  the self-preserving term $\epsilon\mathcal{S}$ in our model, we conduct ablation experiments on different datasets. 
To be specific, we set the hyper-parameter $\epsilon$ to 0 and compare it with the experimental results of adding $\epsilon\mathcal{S}$. The experimental results are shown in Table \ref{tab:ablation}. It demonstrates that adding $\epsilon \mathcal{S}$ term can improve the final learning performance which shows the benefit of this term.
Also, in order to explore more possibilities of the model, we try to replace $A$ in Eq.(5) with the combination of $\frac{1}{2}(\widetilde{A}+\alpha)$ and conduct experiments on two datasets.  The experimental results are shown in Table  \ref{tab:edgeatt}. We can note that the combination way can return feasible results but generally does not bring obvious better performance. }
\begin{table}[htbp]
		\centering
        \caption{Learning performance of ET-GCN across different network depths on Cora dataset.}
        \label{tab:depth}
		\begin{tabular}{c|c|c|c|c|c}
\hline
            Layer&2&3&4&5&6\\\hline
            Accuracy&81.3&80.4&79.9&80.1&80.2\\
            \hline
		\end{tabular}
\end{table}
\begin{table}[htbp]
		\centering
        \caption{Learning performance of ET-GCN across different hidden units on Cora dataset.}
        \label{tab:hidden}
		\begin{tabular}{c|c|c|c|c|c}
\hline
			Hidden Unit&4&8&16&24&32\\\hline
            Accuracy&79.8&81.3&80.8&80.5&80.8\\
            \hline
		\end{tabular}
\end{table}
\begin{table}[htbp]
		\centering
        \caption{Ablation experiments with the adding $\epsilon\mathcal{S}$.}
        \label{tab:ablation}
        \begin{tabular}{c|c|c|c|c}
         \hline
        	\multirow{2}{*}{Dataset} & \multicolumn{2}{c|}{ET-GCN}& \multicolumn{2}{c}{ET-GAT} \\\cline{2-5}
        &$\epsilon$=0&$\epsilon$=0.2&$\epsilon$=0 &$\epsilon$=0.2\\  \hline
        Cora&74.9$\pm$0.7  &75.2$\pm$0.9&          74.1$\pm$0.8&74.8$\pm$0.4\\
        Citeseer&62.5$\pm$1.1&62.6$\pm$0.8&        62.6$\pm$0.8&62.3$\pm$0.6\\
        ogbn-arxiv&71.88$\pm$0.25 &72.01$\pm$0.25&   72.04$\pm$0.22&72.13$\pm$0.18\\
        
         \hline
        \end{tabular}
\end{table}
\begin{table}[htbp]
		\centering
        \caption{Comparison results of different edge weight.}
        \label{tab:edgeatt}
		\begin{tabular}{c|c|c|c}
\hline
            Dataset&$\widetilde{A}$&$\alpha$&$\widetilde{A}+\alpha$
            \\\hline
            Cora&74.8&74.6&75.0\\\hline
            ogbn-arxiv&72.14&72.19&72.16\\
            \hline
		\end{tabular}
\end{table}
\section{Conclusion}

In this paper, we propose a novel GC-like  convolution operation, named Tensor Product Graph Convolution (TPGC) for edge feature representation and embedding.
The core idea behind TPGC is to compute contextual representations for graph edges by adapting tensor contraction layer representation and tensor product graph diffusion mechanism.
Using the proposed TPGC, we propose a unified network architecture, named Edge Tensor-Graph Convolutional Network (ET-GCN) and ET-GAT for the general
graph learning problem with both node and edge features.
Experimental results on several graph learning tasks validate the effectiveness and benefits of TPGC and ET-GCN/ET-GAT.

\ifCLASSOPTIONcaptionsoff
  \newpage
\fi

\bibliographystyle{IEEEtran}
\bibliography{mybibfile}
\end{document}